\pdfoutput=1

\documentclass[11pt]{article}

\usepackage[preprint]{acl}

\usepackage{times}
\usepackage{latexsym}

\usepackage[T1]{fontenc}

\usepackage[utf8]{inputenc}

\usepackage{microtype}

\usepackage{inconsolata}

\usepackage{graphicx}

\usepackage{tabularx}
\usepackage{float}
\usepackage{arabtex}
\usepackage{utf8}
\setcode{utf8}

\usepackage{longtable}
\usepackage{enumitem}
\usepackage{amsmath}
\usepackage{multirow}
\usepackage{booktabs}
\usepackage{listings}
\usepackage{xcolor}
\colorlet{punct}{red!60!black}
\definecolor{background}{HTML}{EEEEEE}
\definecolor{delim}{RGB}{20,105,176}
\colorlet{numb}{magenta!60!black}
\lstdefinelanguage{json}{
    basicstyle=\ttfamily\fontfamily{fvm}\selectfont,
    showstringspaces=false,
    breaklines=true,
    frame=lines,
    backgroundcolor=\color{background},
    literate=
     *{0}{{{\color{numb}0}}}{1}
      {1}{{{\color{numb}1}}}{1}
      {2}{{{\color{numb}2}}}{1}
      {3}{{{\color{numb}3}}}{1}
      {4}{{{\color{numb}4}}}{1}
      {5}{{{\color{numb}5}}}{1}
      {6}{{{\color{numb}6}}}{1}
      {7}{{{\color{numb}7}}}{1}
      {8}{{{\color{numb}8}}}{1}
      {9}{{{\color{numb}9}}}{1}
      {:}{{{\color{punct}{:}}}}{1}
      {,}{{{\color{punct}{,}}}}{1}
      {\{}{{{\color{delim}{\{}}}}{1}
      {\}}{{{\color{delim}{\}}}}}{1}
      {[}{{{\color{delim}{[}}}}{1}
      {]}{{{\color{delim}{]}}}}{1},
}
\title{Atlas-Chat: Adapting Large Language Models for Low-Resource Moroccan Arabic Dialect}

\author{
 \textbf{Guokan Shang\textsuperscript{1}$^\dagger$},
 \textbf{Hadi Abdine\textsuperscript{1}$^\dagger$},
 \textbf{Yousef Khoubrane\textsuperscript{2,3}$^\dagger$},\\
 \textbf{Amr Mohamed\textsuperscript{1}},
 \textbf{Yassine Abbahaddou\textsuperscript{6}},
 \textbf{Sofiane Ennadir\textsuperscript{4}},
 \textbf{Imane Momayiz\textsuperscript{5}},
\\
 \textbf{Xuguang Ren\textsuperscript{1}},
 \textbf{Eric Moulines\textsuperscript{1,6}},
 \textbf{Preslav Nakov\textsuperscript{1}},
 \textbf{Michalis Vazirgiannis\textsuperscript{1,6}},
 \textbf{Eric Xing\textsuperscript{1}}
\\
\\
 \textsuperscript{1}MBZUAI,
 \textsuperscript{2}EMINES-UM6P,
 \textsuperscript{3}LINAGORA,
 \textsuperscript{4}KTH,
 \textsuperscript{5}AtlasIA,
 \textsuperscript{6}Ecole Polytechnique
}
\begin{document}
\maketitle

\def\thefootnote{$\dagger$}\footnotetext{These authors contributed equally.}\def\thefootnote{\arabic{footnote}}
\def\thefootnote{$\dagger$}\footnotetext{Correspondence: \href{mailto:guokan.shang@mbzuai.ac.ae}{guokan.shang@mbzuai.ac.ae}}\def\thefootnote{\arabic{footnote}}

\begin{abstract}
    We introduce Atlas-Chat, the first-ever collection of LLMs specifically developed for dialectal Arabic. 
    Focusing on Moroccan Arabic, also known as Darija, we construct our instruction dataset by consolidating existing Darija language resources, creating novel datasets both manually and synthetically, and translating English instructions with stringent quality control.
    \texttt{Atlas-Chat-2B}, \texttt{9B}\footnote{\url{https://hf.co/MBZUAI-Paris/Atlas-Chat-9B}}, and \texttt{27B} models, fine-tuned on the dataset, exhibit superior ability in following Darija instructions and performing standard NLP tasks.
    Notably, our models outperform both state-of-the-art and Arabic-specialized LLMs like LLaMa, Jais, and AceGPT, e.g., our \texttt{9B} model gains a 13\% performance boost over a larger 13B model on DarijaMMLU, in our newly introduced evaluation suite for Darija covering both discriminative and generative tasks.
    Furthermore, we perform an experimental analysis of various fine-tuning strategies and base model choices to determine optimal configurations.
    All our resources are publicly accessible, and we believe our work offers comprehensive design methodologies of instruction-tuning for low-resource languages, which are often neglected in favor of data-rich languages by contemporary LLMs.
\end{list}    
\end{abstract}

\section{Introduction}
Transformer-based Large Language Models have revolutionized NLP research and beyond, demonstrating exceptional performance in both natural and formal language generation \citep{gunasekar2023textbooks}, and exhibiting advanced reasoning capabilities in arithmetic, symbolic, and logical tasks \citep{hendrycks2020measuring}.	
Despite their success and the frequent release of new, superior open models exemplified by LlaMa \citep{dubey2024llama} and Mistral \citep{jiang2023mistral}, these breakthroughs have been concentrated in a few data-rich languages \citep{ustun-etal-2024-aya}, assuming access to hundreds of billions or even a dozen trillions of tokens for training, often neglecting underrepresented languages.

In this work, we explore the challenges of introducing LLMs for low-resource Dialectal Arabic (DA).
The Arabic language has a rich history and profound cultural significance, featuring an intricate script, extensive lexicon, and complex grammar, making it a unique linguistic entity.
Although interest in developing Arabic-specialized models has recently been growing, notably led by models like Jais \citep{sengupta2023jais}, AceGPT \citep{huang-etal-2024-acegpt}, and ALLaM \citep{bari2024allam}, these efforts primarily focus on bilingualism by balancing English and Modern Standard Arabic (MSA), while often neglecting or excluding DA.
However, MSA differs significantly from DA in terms of morphology, syntax, and other linguistic features. Moreover, various Arabic dialects also differ considerably from one another.	
In fact, Arabic dialects collectively have more native speakers than MSA, as DA serves as the primary mode of communication in daily life across various Arabic-speaking regions \citep{zaidan-callison-burch-2014-arabic}.
This asymmetry is due in large part to the fact that DA poses challenges not encountered with MSA.
Some are related to the lack of essential components for model development---namely, training data, benchmarks, and suitable evaluation metrics---but others stem from the very nature of the linguistic characteristics involved in DA itself more generally.

We take Moroccan Arabic, also known as Darija, as the focus of our work.
Despite being spoken by 40 million people\footnote{\url{https://en.wikipedia.org/wiki/Moroccan_Arabic}}, Darija remains low-resource. This is because MSA is used in official domains in Morocco, while Darija, a blend of MSA, Amazigh, French, and Spanish, is the vernacular widely spoken in daily life.
Although Darija, previously only an oral language, has recently developed a written form through the proliferation of social networks and increased access to technology, it still lacks standardization and established grammatical or syntactic rules due to its recent emergence \citep{gaanoun2024darijabert}.
Moreover, Darija can be represented in two forms: Arabic script or Latin script (also known as Arabizi). For example, the Darija translation of ``How are you?'' can be written as: ``kidayr?'' or ``\<كيداير؟>''.
These challenges underscore the need for models tailored to this linguistic context. 

To the best of our knowledge, we are the first to introduce modern LLMs specifically developed for Moroccan Arabic, as well as for DA in general. 
We first constructed the \texttt{Darija-SFT-Mixture}\footnote{\url{https://hf.co/datasets/MBZUAI-Paris/Darija-SFT-Mixture}} dataset, consisting of 458K instruction samples, by consolidating existing Darija language resources, creating novel datasets both manually and synthetically, and translating English instructions under strict quality control.
We then developed a comprehensive evaluation suite including benchmarks: \texttt{DarijaMMLU}, \texttt{DarijaHellaSwag}, \texttt{DarijaAlpacaEval}, and \texttt{DarijaBench}, to assess LLM capabilities in real-world knowledge, following Darija instructions, and performing traditional NLP tasks such as translation, summarization, and sentiment analysis.	
In the end, Atlas-Chat models\footnote{Inspired by the naming of the ``Jais'' models, UAE’s highest mountain peak. We chose ``Atlas'' to reflect the cultural and geographical significance of the Atlas Mountains that traverse Morocco.}, fine-tuned from the Gemma 2 models \citep{team2024gemma} on our instruction dataset, exhibit superior ability in Darija, surpassing both state-of-the-art and Arabic-specialized LLMs like LLaMa, Jais, and AceGPT, according to automatic metrics and simulated win rates.
Additionally, we conduct an experimental analysis of various fine-tuning strategies and base model choices to determine final configurations.
We provide some examples by chatting with our models in Appendix \ref{app:chat_examples}.
All our resources are publicly accessible, and we believe our work offers comprehensive design methodologies of instruction-tuning for low-resource languages.


\section{Related Work}
In this section, we begin by reviewing LLMs and benchmarks developed for Arabic, followed by an exploration of recent trends in expanding LLMs to low-resource languages.	

\smallskip

\noindent\textbf{Arabic-specialized LLMs}. 
Recent efforts in Arabic-specialized LLMs mainly focus on MSA, the formal written standard across Arabic regions.

\textbf{\textit{Jais}} \citep{sengupta2023jais}, a 13B-parameter model trained on 395B tokens of Arabic, English, and code data. Containing 116B Arabic tokens---25\% of which were translated from English---Jais was designed to enhance performance in both Arabic and English tasks, trained on a mixture of the two languages in a 1:2 ratio. However, this approach may suffer from localization issues.
\textbf{\textit{AceGPT}} \citep{huang-etal-2024-acegpt} aims to address localization issues by pre-training LLaMA 2 \citep{touvron2023llama} 7B and 13B models on 30B and 10B token mixtures, respectively, of Arabic and English data, with the Arabic portion dominating the dataset. The models were then fine-tuned on Arabic instructions and aligned with Arabic values and culture using RLAIF \citep{lee2023rlaif}. They further introduced the Arabic Cultural and Value Alignment dataset, comprising 8,000 yes-no questions.
\textbf{\textit{ALLaM}} \citep{bari2024allam} demonstrated that second-language acquisition can steer the model towards a new language without catastrophic forgetting, even with random initialization of weights. They hypothesize that low-resource languages are diluted in large volumes of high-resource languages, and pre-train a 7B model from scratch on 4T English tokens, followed by training on a 1.2T mixture of Arabic and English.

Regarding \textbf{\textit{Darija}}, DarijaBERT \citep{gaanoun2024darijabert} is currently the only ``LLM'' dedicated to the Moroccan Arabic dialect. The model was trained on $\sim$100M tokens. However, DarijaBERT is encoder-only, and no decoder-only models have been developed for Darija.

\smallskip

\noindent\textbf{Arabic benchmarks for LLMs}. 
Several benchmarks have been created for various tasks and domains to evaluate the Arabic capabilities of LLMs.

\textbf{\textit{ArabicMMLU}} \citep{koto-etal-2024-arabicmmlu} is an Arabic adaptation of the original MMLU benchmark \citep{hendrycks2020measuring}, consisting of 14K multiple-choice questions across 40 tasks in MSA. The benchmark covers a wide range of subjects, including history, mathematics, science, and linguistics, reflecting educational levels from eight different countries.	
\textbf{\textit{LAraBench}} \citep{abdelali-etal-2024-larabench}, a benchmark designed for evaluating MSA LLMs on several practical NLP tasks, such as sentiment analysis, named entity recognition, and machine translation, spanning 33 tasks across 61 datasets encompassing $\sim296$ data points.
The Open Arabic LLM Leaderboard (\textbf{\textit{OALL}})\footnote{\url{https://hf.co/blog/leaderboard-arabic}} aggregates various native and translated Arabic benchmarks to evaluate models' performance across tasks such as reading comprehension, reasoning, and more.

\smallskip

\noindent\textbf{LLMs for Low-resource languages}.
Very recently, the LLM development community has begun to focus on low-resource languages.

Multilingual \textbf{\textit{Aya}} model \citep{ustun-etal-2024-aya} was developed by instruction-tuning mT5 \citep{xue2020mt5}, a 13B encoder-decoder model pre-trained on 1T tokens across 101 languages. Of these, 51 are low-resource languages, including Hausa, Icelandic, Luxembourgish, and etc.	
Other efforts include \textbf{\textit{InkubaLM}} \citep{tonja2024inkubalm}, a 0.4B model pre-trained from scratch on 2.4B tokens from five low-resource African languages---Hausa, Yoruba, Swahili, isiZulu, and isiXhosa---along with English and French, then fine-tuned to follow instructions on several tasks. Similarly, \citet{tao2024unlocking} explored two language adaptation strategies: continual pre-training followed by fine-tuning and model merging. Their experiments focused on seven low-resource languages---Tamil, Telugu, Odia, Bengali, Tibetan, Uyghur, and Mongolian---using datasets ranging from 1 to 20B tokens per language. Another line of research targets a subcategory of main languages with limited resources, such as the \textbf{\textit{Claire}} model \citep{hunter2023claire, louradour-etal-2024-claire}, dedicated to spontaneous French dialogue.

\smallskip

Despite advancements, little attention has been given to developing LLMs and benchmarks for DA.


\begin{table*}[ht]
\centering
\renewcommand{\arraystretch}{1.3} 
\setlength{\tabcolsep}{3pt}
\footnotesize
\begin{tabular}{p{3.5cm} p{1.5cm} p{3.5cm} p{7cm}}
\hline
\textbf{Subset} & \textbf{\# Samples} & \textbf{Source} & \textbf{Description} \\
\hline
§ \ref{sec:data:translation} \textbf{Translation} & 85,662 & DODa-10K, FLORES+, MADAR, NLLB-Seed & Darja to English, French, MSA and vice-versa \\
§ \ref{sec:data:translation} \textbf{Transliteration} & 16,920 & DODa-10K & Darija in Arabic Script $\leftrightarrow$ Latin Script \\
§ \ref{sec:data:sentiment} \textbf{Sentiment Analysis} & 86,212 & MSAC, MSDA, MAC \newline ElecMorocco2016, MYC & Sentences labeled as Positive, Negative, and Neutral \\
§ \ref{sec:data:summarization} \textbf{Summarization} & 16,756 & MArSum & Article titles as summaries \\
§ \ref{sec:data:mw_qa} \textbf{MW-QA} & 30,555 & Wikipedia & Synthetic dataset from Moroccan Wikipedia pages \\
§ \ref{sec:data:msm_mg} \textbf{MSM-MG} & 11,808 & Twitter/X, YouTube Comments & Synthetic dataset from Tweets and YouTube comments \\
§ \ref{sec:data:story} \textbf{Story Completion} & 48,983 & 9esa.com & Stories converted to a dataset with part of the story as a prompt and the continuation as a response \\
§ \ref{sec:data:english_tulu} ~~~\textbf{TÜLU-Darija} & 161,259 & TÜLU-V2-Mix & Translated TÜLU-V2-Mix after filtering \\
§ \ref{app:details:hard_coded} \textbf{Hard Coded} & 130 & Manual Annotation & Prompts ensuring the model correctly answers identity/creator-related questions \\
\hline
\end{tabular}
\caption{Composition of our Darija-SFT-Mixture instruction-tuning dataset.}
\label{tab:dataset_info}
\end{table*}

\section{Data Overview}
In developing Atlas-Chat, we chose to use instruction-tuning on a base model rather than training from scratch. This decision was primarily driven by the fact that training an LLM from the ground up requires extensive data, which is not readily available for Darija, a low-resource dialect. For the same reason, our training process does not include the additional continual pre-training phase typically seen in many language adaptation efforts.	
However, to mitigate this limitation, we designed a synthetic instruction dataset (see Section \ref{sec:data:story}) that, to some extent, mimics the next-word prediction task over a relatively longer context, typically performed during (continual) pre-training.

Moreover, recent studies show that multilingual LLMs often exhibit a bias toward internally solving tasks in English, even when trained on multiple languages \citep{zhao2024large}, and perform best with English prompts, followed by mixed prompts, while non-English prompts significantly underperform \citep{ kmainasi2024native}.	
This observation led us to limit the scope of our work to a monolingual LLM, making Atlas-Chat \textbf{Darija-centric}.
We focus on developing a model that accurately understands prompts written in Darija, generates Darija content, respects its cultural context, and remains accessible and adaptable for native speakers.

Therefore, we directed our efforts towards creating an extensive and diverse Darija dataset for instruction-tuning. Table \ref{tab:dataset_info} summarizes the composition of our Darija-SFT-Mixture dataset. 
We employed a multifaceted approach to data preparation. 
\textit{First}, we reviewed previous research in Darija NLP and collected the majority of available native Darija datasets that met our quality standards.
The data selection rule established by native speakers was as follows: if the data is a mix of Darija with some MSA, it is acceptable; if it is mixed with other dialects, it is not.
In total, ten datasets covering tasks such as translation, summarization, and sentiment analysis were selected.
\textit{Second}, we synthesized high-quality instruction data using advanced proprietary models, drawing on sources such as Wikipedia pages, social media posts, and stories written in Darija.
We then converted the native and synthetic datasets into training instructions using templates, with 80\% formatted as zero-shot, 10\% as few-shot \citep{longpre2023flan}, and 10\% as multi-turn samples.	
\textit{Third}, we translated high-quality English instruction datasets into Darija with stringent quality control to expand the range of scenarios, domains, and tasks covered by our dataset.
By combining these different sources, we aimed to enhance the model's ability to understand and generate Darija across various contexts.

\section{Native Darija Instruction Datasets}
\label{sec:native_darija_datasets}

\subsection{Machine Translation}
\label{sec:data:translation}
We collected four datasets containing sentence translations between Darija, MSA, English, and French.	
These datasets were then converted into training instructions using the templates provided in Appendix \ref{app:instruction_templates:machine_translation}.
Since our model is Darija-centric, we consider six translation directions: Darija to English, French, MSA, and vice versa. All instructions are written in Darija for each case.	

\smallskip

\noindent\textbf{DODa-10K}\footnote{\url{https://hf.co/datasets/MBZUAI-Paris/DoDa-10K}}. The Darija Open Dataset (DODa) \citep{outchakoucht2021moroccan,outchakoucht2024evolution}\footnote{\url{https://github.com/darija-open-dataset}} is an open-source collaborative project for collecting Darija language resource, including lexicons in semantic and syntactic categories, Darija-English parallel corpus, and etc. Darija is represented in Latin script, as well as in automatically converted Arabic script. 
We augmented the first 10K examples of the parallel corpus, with MSA and French translated from the English text, by leveraging GPT-4. 
The final DODa-10K dataset includes translation quintuples between \textit{Darija} (in both Arabic and Latin scripts), \textit{MSA}, \textit{English}, and \textit{French}.
The dataset was then extensively reviewed by groups of native speakers to ensure the quality.

In addition to translation, to enhance the model's ability to convert between Darija in Arabic and Latin scripts (also known as the \textbf{\textit{transliteration}} task), we transformed 10K parallel forms into instructions using templates found in Appendix \ref{app:instruction_templates:transliteration}.

\smallskip

\noindent\textbf{MADAR} \citep{bouamor-etal-2018-madar}\footnote{\url{https://sites.google.com/nyu.edu/madar}}. The Multi-Arabic Dialect Applications and Resources (MADAR) corpus is a collection of parallel sentences covering the dialects of 25 Arab cities, built upon the Basic Traveling Expression Corpus \citep{takezawa-etal-2007-multilingual}. We select the dialect of Rabat city as \textit{Darija} translation, along with \textit{MSA}, resulting in 12K sentence pairs.
The split corpus-6-test-corpus-26-test is reserved for the evaluation.

\smallskip

\noindent\textbf{NLLB-Seed} \citep{maillard-etal-2023-small}\footnote{\url{https://github.com/openlanguagedata/seed}}. The Seed machine translation dataset contains 6K sentences sampled from English Wikipedia and translated into 39 low-resource languages. We extract the \textit{Darija} and \textit{English} pairs.

\smallskip

\noindent\textbf{FLORES+}\footnote{\url{https://github.com/openlanguagedata/flores}}. Built upon FLORES-200 \citep{costa2022no}, this corpus is specifically designed to support multilingual research and evaluation. The English sentences were sampled in equal amounts from Wikinews, Wikijunior (a collection of age-appropriate non-fiction books), and Wikivoyage. These were then translated into other languages.
For each language, the dataset has 997 sentences for the dev split and 1012 sentences for the devtest split.
We selected those in \textit{Darija}, \textit{MSA}, \textit{English}, and \textit{French}.
Dev is severed as training, while devtest for the evaluation.

\subsection{Sentiment Analysis}
\label{sec:data:sentiment}
We collected five datasets for sentiment analysis, whose content is primarily sourced from social networks.
Two datasets come with three labels (positive, negative, and neutral), while the other three have two labels (positive and negative).
These datasets were then transformed into training instructions using templates from Appendix \ref{app:instruction_templates:sentiment_analysis}.

\smallskip

\noindent\textbf{MSDA} \citep{boujou2021open}\footnote{\url{https://cc.um6p.ma/cc_datasets}}. It is an open dataset for sentiment analysis, designed to support research in NLP for Arabic dialects and social media.
The dataset includes 52K tweets in Darija, categorized into three labels: \textit{positive}, \textit{neural}, or \textit{negative}.
The tweets are preprocessed, and emojis are retained because they play a significant role in expressing sentiment.
Labels are annotated semi-automatically and bootstrapped with human intervention.

\smallskip

\noindent\textbf{MSAC} \citep{oussous2018improving,oussous2020asa}\footnote{\url{https://github.com/ososs/Arabic-Sentiment-Analysis-corpus}}. The Moroccan Sentiment Analysis Corpus (MSAC) is a manually prepared dataset consisting of reviewers’ opinions for Hespress\footnote{\url{https://www.hespress.com}} articles, and a collection of Arabic comments from Facebook, Twitter and YouTube. It includes content in both MSA and Darija, consisting of 2K sentences labeled as \textit{positive} or \textit{negative} in equal proportions.

\smallskip

\noindent\textbf{ElecMorocco2016} \citep{elouardighi2017collecting}\footnote{\url{https://github.com/sentiprojects/ElecMorocco2016}}. 
The 2016 Moroccan elections (ElecMorocco2016) is a sentiment analysis dataset comprising 10K Facebook comments about Moroccan’s
legislative elections held on October 7, 2016. Each comment is labeled as either \textit{positive} or \textit{negative}. The comments are written in Darija and MSA.

\smallskip

\noindent\textbf{MYC} \citep{jbel2024sentiment}\footnote{\url{https://github.com/MouadJb/MYC}}.
The Moroccan Youtube Corpus (MYC) is a sentiment analysis dataset of YouTube comments collected from Moroccan channels covering various topics.
The dataset prioritizes variety over size, with 20K manually labeled samples, evenly divided between \textit{positive} and \textit{negative}.	
Notably, the 20K comments are equally balanced between Arabic script and Latin script.	

\smallskip

\noindent\textbf{MAC} \citep{garouani2021mac}\footnote{\url{https://github.com/LeMGarouani/MAC}}. The Moroccan Arabic Corpus (MAC) is a free, large-scale Darija corpus for sentiment analysis, consisting of 18K manually labeled tweets categorized as \textit{positive}, \textit{neutral}, \textit{negative}, or mixed. Only 643 tweets are labeled as mixed, so we filtered them out.	

\subsection{Automatic Summarization}
\label{sec:data:summarization}
We found only one dataset for summarization. The documents and summaries were converted into instructions using the template in Appendix \ref{app:instruction_templates:automatic_summarization}.

\smallskip

\noindent\textbf{MArSum} \citep{gaanoun2022automatic}\footnote{\url{https://github.com/KamelGaanoun/MoroccanSummarization}}.
The Moroccan Articles Summarization dataset (MArSum) contains 19K news articles written in Darija, along with their titles. The articles were crawled from  Goud.ma\footnote{\url{http://www.goud.ma/}}. While some content includes MSA, all titles are written in Darija. Since the articles are relatively concise and the titles are sufficiently informative, the titles are considered as summaries. The average length of the titles is 14.6 words.

\section{Synthetic Darija Instruction Datasets}

\subsection{MoroccanWikipedia-QA}
\label{sec:data:mw_qa}
\textbf{MW-QA}\footnote{\url{https://hf.co/datasets/MBZUAI-Paris/MoroccanWikipedia-QA}} is a dataset derived from Moroccan Wikipedia dump\footnote{\url{https://dumps.wikimedia.org/arywiki/latest/}}, developed in our work to enhance the models' question-answering (QA) capability. The dataset is divided into four tasks: Open QA (8\%), Multiple-Choice QA (40\%) (MMLU-alike), Extractive QA (10\%), and Multiple-Choice Extractive QA (42\%) (Belebele-alike), with each percentage reflecting the proportion of Wikipedia pages used for the respective task. 
The latter two tasks provide context along with the questions, whereas the former two do not. In Open QA and Extractive QA, answers are provided in sentence form. In the multiple-choice tasks, four answer options are presented, with the index of the correct option serving as the answer.
The distribution of correct answers (e.g., A, B, C, D) are balanced.
The QAs were converted into instructions with the template in Appendix \ref{app:instruction_templates:moroccanwikipedia_QA}.

The dataset generation involved providing each Wikipedia page to Claude 3.5 Sonnet\footnote{\url{https://www.anthropic.com/news/claude-3-5-sonnet}} and prompting it to generate QA pairs tailored to the four task categories. The prompts followed a one-shot or two-shot format to ensure that output adhered to the desired structure. For the extractive tasks, rather than splitting the page into paragraphs—an approach that risked losing contextual meaning—we opted to present the entire page to Claude. The model was instructed to first extract a meaningful passage from the page and then generate a QA pair based on the content of that passage. Also, the model was directed to ensure that the extracted passages were long, self-contained, and did not lose meaning when removed from their original context.

A total of 8,730 pages were collected and pre-processed by removing scraping errors. Among these pages, some followed a uniform structure, typically consisting of a brief description of a village or community followed by statistical data (e.g., literacy rates and unemployment figures). Given that these statistical sections could become meaningless when extracted from their context, they were allocated to non-extractive tasks, which could still utilize the statistical information to enrich the fine-tuned model's knowledge base.

The final distribution of QA pairs is as follows: 15.7\% Open QA, 43.1\% Multiple-Choice QA, 6.9\% Extractive QA, and 34.3\% Multiple-Choice Extractive QA. These percentages differ from the initial page distribution because Claude generated varying numbers of samples for each task. For example, the average number of samples generated for Open QA is 7.73, while for Extractive QA, it is 2.72.

\subsection{MoroccanSocialMedia-MultiGen}
\label{sec:data:msm_mg}
\textbf{MSM-MG}\footnote{\url{https://hf.co/datasets/MBZUAI-Paris/MoroccanSocialMedia-MultiGen}}, a dataset introduced as part of this work, comprises 12,973 pairs of native Darija social media posts (tweets and YouTube comments) and their synthetic counterparts, covering various NLP tasks. The pairs were converted into instructions using the template provided in Appendix \ref{app:instruction_templates:moroccansocialmedia_multigen}.

The synthetic generations are created based on six specific tasks: \textit{Continuation}, \textit{Reply}, \textit{Summarization}, \textit{Rephrasing}, \textit{Explanation}, and \textit{Safe Response}, by prompting Claude 3.5 Sonnet to respectively consider the original post as incomplete and continue it, reply to it, summarize its content, rephrase it, explain its topic, and respond safely to potentially offensive content. 9,754 Tweets were employed for the first five tasks, while 3,219 YouTube comments were utilized for the last task.
The posts were collected from three sources:\\ \noindent\textit{QADI} \citep{abdelali-etal-2021-qadi}\footnote{\url{https://github.com/qcri/QADI}}: From this Arabic dialect identification dataset, 12,813 Moroccan tweets were initially sampled. After a thorough review by native speakers, tweets that were no longer accessible or contained non-Darija Arabic dialects were filtered out, resulting in 6,362 valid tweets.
    
\noindent\textit{Twitter API}: 4,226 tweets were gathered directly from the Twitter API by searching for Darija-specific keywords. The DarijaBERT work identified 31 keywords exclusive to Darija, but upon review, five were found to also exist in other Arabic dialects and were excluded. The remaining 26 keywords can be found in Appendix \ref{app:details:keywords}.

\noindent\textit{OMCD} \citep{essefar2023omcd}\footnote{\url{https://github.com/kabilessefar/OMCD-Offensive-Moroccan-Comments-Dataset}}: This is a dataset for offensive content identification collected from Moroccan YouTube comments. For our work, only comments labeled as offensive from the training split were selected. We then utilized these offensive comments for the generation of synthetic safe responses specifically. 

\subsection{DarijaStory-Completion}
\label{sec:data:story}
To mitigate the limitation of performing only instruction-tuning for language adaptation without the typical continual pre-training phase---due to the lack of sufficient amount of Darija pre-training data---we designed a synthetic story completion dataset, aiming to enhance the next-word prediction capability in Darija for our models over a relatively longer context.	
First, we collected 4,392 long stories from 9esa\footnote{\url{https://www.9esa.com}}, a website featuring a rich collection of various stories entirely written in Darija. 
We denote this dataset as DarijaStory\footnote{\url{https://hf.co/datasets/MBZUAI-Paris/DarijaStory}}.
The scraped stories were then divided into segments of approximately 2,048 tokens, adhering to the base model tokenizer's vocabulary.	
The segments were further divided into two parts of varying lengths: the beginning part and the ending part to be completed.	
For the two segmentation steps above, the split point is preferably placed at line breaks.	
Finally, the pairs were converted into instructions using the template provided in Appendix \ref{app:instruction_templates:darijastory_9esa}.

\section{Translated English Instruction Datasets}
\label{sec:data:english_tulu}

Finally, we broadened our instruction-tuning data by translating English datasets into Darija, to cover a wider array of scenarios, domains, and tasks.

We began by reviewing the most widely used datasets for fine-tuning state-of-the-art models to ensure that our translation efforts would lead to meaningful improvements.	
After careful consideration, we decided to focus on the \textbf{TÜLU-V2-mix} \citep{ivison2023camels}\footnote{\url{https://hf.co/datasets/allenai/tulu-v2-sft-mixture}} dataset for several reasons.	
It offers a comprehensive dataset composition, including samples from some of the most widely used datasets, such as FLAN and ShareGPT, for fine-tuning state-of-the-art models.
Appendix \ref{app:tulu:composition} presents descriptions of each of these datasets and describes how the subset was sampled.
The dataset mixture was meticulously designed based on ablation studies of both human-annotated and AI-generated data, with a focus on complexity and diversity.	
Models fine-tuned on it showed significant improvements in overall performance on key benchmarks compared to those trained on individual datasets.
We adopted the user-assistant message format from TÜLU-V2-mix (see Appendix \ref{app:tulu:format}) to structure our entire Darija-SFT-Mixture dataset.

To ensure quality, we first filtered out instructions from TÜLU-V2-mix that are either inappropriate for typical Darija speakers or could lose meaning or coherence when translated, such as scientific content, translation tasks, and non-English samples.	
We then experimented with several open-source and closed-source models for English-to-Darija translation, including NLLB \citep{costa2022no}, GPT, and others.	
Our results showed that closed-source models consistently outperformed open-source alternatives, with Claude 3.5 Sonnet emerging as our final choice.
Finally, we implemented several post-processing measures to correct errors introduced by the automatic translation.	
All details are provided in Appendix \ref{app:tulu:translation}.

\section{Training Details}
In this section, we outline the training details and present the experimental analysis of various fine-tuning strategies and base model choices that informed our final settings.	

\smallskip

\noindent\textbf{Base model selection}.
Initially, we considered the two Arabic models: Jais and AceGPT (as ALLaM is not open-weights).
Later, we included Gemma 2 based on positive feedback from Arabic LLM community, as it can serve as a strong starting point for Arabic fine-tuning tasks.	
We also compared the performance differences between fine-tuning on an instruction-tuned model and a base model.
Our results indicate that \textit{continual fine-tuning} of instruction-tuned Gemma 2 models (Gemma-2-2B-It, 9B-It\footnote{\url{https://hf.co/google/gemma-2-9b-it}}, and 27B-It) yields significantly higher scores than other settings on our dataset.

\smallskip

\noindent\textbf{Training framework}.
We also investigated the performance differences between full fine-tuning and parameter-efficient approaches \citep{han2024parameter}.
Results indicate that the latter, with Low-Rank Adaptation (LoRA) \citep{hu2021lora}, proved to be more effective, whereas full fine-tuning resulted in catastrophic forgetting \citep{french1999catastrophic}.	
This is supported by the recent work of \citet{biderman2024lora}, that shows LoRA exhibits a desirable form of regularization: it better maintains the base model's performance on tasks outside the target domain, and it also helps maintain more diverse generations.

\smallskip

\noindent\textbf{Hyperparameters}.
LoRA was set with rank 256 and alpha 128. 
We run the training for 3 epochs, and set the learning rate to 5e-5 with warmup ratio of 3\%, and per\_device\_train\_batch\_size to 4, with gradients accumulated over 4 steps.
The maximum input context length was configured to 2048.
We used bfloat16 to optimize training speed.
The loss is computed only on the responses, not on the prompts of instructions.
The Atlas-Chat models were trained on 8 Nvidia A100 80 GB GPUs in parallel, utilizing FSDP strategy on AWS SageMaker. 

\section{Evaluation Benchmarks}
To evaluate LLM performance in Darija, we developed a comprehensive suite that includes benchmarks such as DarijaMMLU, DarijaHellaSwag, DarijaAlpacaEval, and DarijaBench.	
Additionally, we evaluated using an existing benchmark, Belebele.
All our custom benchmarks are integrated into a fork\footnote{\url{https://github.com/MBZUAI-Paris/lm-evaluation-harness-atlas-chat}} of the LM-Evaluation-Harness repository \citep{eval-harness} to ensure reproducibility and foster future model comparison.	

\smallskip

\noindent\textbf{DarijaMMLU}\footnote{\url{https://hf.co/datasets/MBZUAI-Paris/DarijaMMLU}}. It is constructed by translating two major benchmarks into Darija from English and MSA: Massive Multitask Language Understanding (MMLU) \citep{hendrycks2020measuring}\footnote{\url{https://hf.co/datasets/cais/mmlu}} and ArabicMMLU \citep{koto-etal-2024-arabicmmlu}\footnote{\url{https://hf.co/datasets/MBZUAI/ArabicMMLU}},
whose subsets that were either too technical (beyond typical user needs) or culturally inappropriate for the Moroccan context were excluded. 
The remaining samples were translated into Darija using Claude 3.5 Sonnet. The benchmark consists of 22,027 multiple-choice questions, with the number of choices ranging from 2 to 5. The  subsets we selected are listed in \ref{app:details:mmlu}. 

\smallskip

\noindent\textbf{DarijaHellaSwag}\footnote{\url{https://hf.co/datasets/MBZUAI-Paris/DarijaHellaSwag}}.
HellaSwag\footnote{\url{https://hf.co/datasets/Rowan/hellaswag}} \citep{zellers-etal-2019-hellaswag} is a multiple-choice dataset designed to evaluate machine reading comprehension and commonsense reasoning. It presents complex scenarios where models must select the most plausible continuation of a passage from four options, challenging nuanced language understanding and contextual inference. Using Claude 3.5 Sonnet, We translated the HellaSwag validation set into Darija.

\smallskip

\noindent\textbf{Belebele\_Ary}.
Belebele \citep{bandarkar-etal-2024-belebele}\footnote{\url{https://hf.co/datasets/facebook/belebele}} is a multiple-choice machine reading comprehension dataset designed to evaluate both monolingual and multilingual models across 122 languages. Each question is paired with a brief passage and offers four multiple-choice answers. For our work, we specifically used the Ary\_Arab (indicating Moroccan Arabic) subset of Belebele.


\smallskip

\noindent\textbf{DarijaAlpacaEval}\footnote{\url{https://hf.co/datasets/MBZUAI-Paris/DarijaAlpacaEval}}. Claude 3.5 Sonnet was prompted to translate and culturally adapt the AlpacaEval dataset \cite{alpaca_eval}
into Darija, to evaluate the instruction-following capabilities and cultural alignment of LLMs in Darija. The dataset consists of 805 instructions, focusing on culturally relevant content tailored to the Moroccan context. More details about the dataset creation and evaluation method can be found in Appendix \ref{app:details:alpaca}.

\smallskip

\noindent\textbf{DarijaBench}\footnote{\url{https://hf.co/datasets/MBZUAI-Paris/DarijaBench}}.
In addition to the above benchmarks, we evaluated with the test sets from the native Darija datasets (see Section \ref{sec:native_darija_datasets}). Typically, 10\% of each subset is reserved for testing, unless the original source provides a pre-defined separate test set. The combined test sets, referred to as DarijaBench, encompass three tasks: Translation, Sentiment Analysis, and Summarization.

\begin{table*}[ht]
\centering
\renewcommand{\arraystretch}{1.2}
\setlength{\tabcolsep}{5pt}
\resizebox{\textwidth}{!}{%
\begin{tabular}{p{3.2cm} cccccccccccccccc}
\toprule
\multirow{2}{*}{\textbf{\small Base Model}} &
\multicolumn{2}{c}{\shortstack{\textbf{\small DarijaMMLU}}} &
\multicolumn{2}{c}{\shortstack{ \textbf{\small DarijaHellaSwag}}}&
\multicolumn{2}{c}{\shortstack{\textbf{\small Belebele\_Ary}}} & 
\multirow{2}{*}{\shortstack{\textbf{\small Darija} \\ \textbf{\small AlpacaEval}}}& 
\multirow{2}{*}{\shortstack{\textbf{\small Sentiment} \\ \textbf{\small Analysis}}} & \multicolumn{3}{c}{\textbf{\small Translation (DODa-10K)}} &
\multicolumn{5}{c}{\textbf{\small Summarization (MArSum)}} \\
\cmidrule(lr){2-3} \cmidrule(lr){4-5} \cmidrule(lr){6-7} \cmidrule(lr){10-12} \cmidrule(lr){13-17}& 
\footnotesize 0-shot & \footnotesize  3-shot & \footnotesize 0-shot & \footnotesize  10-shot & \footnotesize 0-shot & \footnotesize  5-shot &  & & 
\footnotesize chrF & \footnotesize BLEU & \footnotesize \shortstack{\small BERTScore} &
\footnotesize chrF & \footnotesize ROUGE-1 & \footnotesize ROUGE-L & \footnotesize \shortstack{\small BERTScore} &
\footnotesize \shortstack{LLM Judge} \\
\midrule
\small \textbf{Llama-3.2-1B-Instruct}  & 27.66              & 30.79            & 26.88             & 27.03 & 28.89             & 24.00 & 23.57             & 46.27 & 5.95 & 0.07 & 37.45 & 27.78 & 7.35 & 7.18 & 38.32 & 8.23 \\
\small \textbf{Jais-family-1.3B-chat}  & 35.39              & 31.24            & 27.71             & 27.25 & 38.89             & 37.44 & 35.56             & 44.82 & 6.01 & 0.12 & 39.17 & 20.56 & 6.85 & 6.72 & 35.77 & 0.50 \\
\small \textbf{Gemma-2-2B-It}          & 28.59              & 38.22            & 27.72             & 27.65 & 25.22             & 40.67 & 58.67             & 53.38 & 3.58 & 0.07 & 35.31 & 0.48 & 0.49 & 0.48 & 24.44 & 6.79 \\
\small \textbf{Jais-family-2.7B-chat}  & 37.58              & 31.76            & 29.10             & 28.32 & 45.00             & 38.67 & 52.97             & 51.67 & 7.51 & 0.26 & 39.80 & 20.63 & 7.74 & 7.60 & 36.38 & 0.89 \\
\small \textbf{Llama-3.2-3B-Instruct}  & 32.60              & 31.17            & 28.33             & 28.26 & 38.00             & 40.77 & 47.62             & 49.20 & 13.67 & 0.62 & 43.78 & 27.56 & 8.16 & 8.09 & 38.56 & 8.23\\
\small \textbf{Jais-family-6.7B-chat}  & 39.96              & 33.42            & 32.64             & 32.64 & 51.22             & 46.67 & 65.18             & 56.93 & 11.81 & 0.71 & 45.80 & 22.12 & 7.98 & 7.82 & 37.10 & 3.02 \\
\small \textbf{Jais-Adapted-7B-chat}   & 39.30              & 39.07            & 29.55             & 29.97 & 43.56             & 30.67 & 61.84             & 52.96 & 9.36 & 0.60 & 45.03 & 23.20 & 7.82 & 7.63 & 36.89 & 2.82 \\
\small \textbf{AceGPT-7B-chat}         & 36.00              & 29.31            & 30.33             & 30.83 & 30.33             & 25.67 & 47.31             & 40.18 & 11.34 & 0.45 & 45.36 & 27.18 & 7.60 & 7.55 & 37.29 & 2.28 \\\hline
\small \textbf{Atlas-Chat-2B}          & 45.01              & 44.43            & 35.04             & 34.55 & 53.33             & 56.67 & 92.31             & \underline{74.01} & \textit{44.86} & \textit{22.76} & \textit{73.72} & \textit{28.80} & 9.00 & 8.88 & \textit{44.71} & \textit{55.22} \\\hline
\small \textbf{Llama-3.1-8B-Instruct}  & 44.14              & 44.75            & 31.40             & 31.94 & 47.22             & 28.56 & 78.08             & 44.17 & 13.82 & 0.84 & 44.62 & 28.66 & \underline{10.20} & \underline{9.93} & 39.37 & 16.14 \\
\small \textbf{Gemma-2-9B-It}          & 35.96              & 56.38            & 33.61             & 35.06 & 31.33             & 69.22 & 90.86             & 59.93 & 15.04 & 0.85 & 48.28 & 25.49 & \textit{9.84} & \textit{9.93} & 39.37 & 13.81 \\
\small \textbf{Jais-family-13B-Chat}   & 45.08              & 41.91            & 33.98             & 33.93 & 58.56             & 48.56 & 69.93             & 41.79 & 11.73 & 0.93 & 45.90 & 22.53 & 7.99 & 9.64 & 38.00 & 1.77 \\
\small \textbf{Jais-Adapted-13B-chat}  & 45.31              & 46.92            & 32.84             & 33.25 & 50.11             & 47.33 & 77.52             & 66.85 & 10.48 & 0.88 & 47.85 & 23.80 & 8.86 & 7.84 & 37.13 & 1.92 \\
\small \textbf{AceGPT-13B-chat}        & 41.05              & 36.55            & 32.19             & 33.05 & 33.11             & 36.78 & 52.79             & 59.60 & 14.22 & 0.69 & 47.97 & 26.83 & 7.92 & 8.63 & 37.67 & 2.80 \\\hline
\small \textbf{Atlas-Chat-9B}          & \underline{58.32}  & \textit{59.31}   & \underline{43.65} & \underline{44.83} & \underline{74.33} & \underline{79.44} & \underline{95.62} & \textbf{81.85} & \underline{50.44} & \underline{27.98} & \underline{76.30} & \underline{32.07} & 9.50 & 9.45 & \underline{47.00} & \underline{59.76} \\\hline
\small \textbf{jais-family-30B-8k-chat}& \textit{51.88}     & 49.27            & 35.61             & 36.77 & \textit{65.67}    & 22.89 & 56.73             & 24.64 & 14.40 & 1.10 & 47.22 & 22.31 & 8.15 & 7.97 & 37.17 & 0.46 \\
\small \textbf{gemma-2-27b-it}         & 36.47              & \underline{59.80}&\textit{37.04}     & \textit{39.38} & 35.78             & \textit{75.56} &\textit{95.07}     & 57.59 & 13.04 & 0.67 & 48.17 & 9.64 & 5.62 & 5.52 & 37.22 & 11.10 \\\hline
\small \textbf{Atlas-Chat-27B}         & \textbf{61.95}     & \textbf{63.30}   & \textbf{48.37}    & \textbf{48.72} & \textbf{75.67}    & \textbf{80.67} & \textbf{96.58}    & \textit{73.00} & \textbf{51.74} & \textbf{29.55} & \textbf{77.03} & \textbf{32.75} & \textbf{10.53} & \textbf{10.42} & \textbf{47.82} & \textbf{60.70} \\
\bottomrule
\end{tabular}
}
\caption{Performance comparison of Atlas-Chat and state-of-the-art models on the evaluation suite with prompts written in \textbf{Darija}. The highest scores are indicated in \textbf{bold}, the second-highest are \underline{underlined}, and the third-highest are in \textit{italic}. Figure \ref{fig:overall-accuracy} shows the average score over all the benchmarks and measures for each model.}
\label{tab:results}
\end{table*}

\section{Results}

\noindent\textbf{Evaluation measures}.
We employed Accuracy to evaluate models on multiple-choice benchmarks, including DarijaMMLU, DarijaHellaSwag, Belebele\_Ary, and the discriminative sentiment analysis task within DarijaBench.
For translation and summarization tasks, we adopted the conventional BLEU \citep{papineni-etal-2002-bleu} and ROUGE-1/L \citep{lin-2004-rouge}, respectively.
However, since these metrics are based on $n$-grams, they are not well-suited for assessing Darija. 
For example, the same word in Darija can be written in multiple ways ("How are you?" = "\<كيدير>" = "\<كيداير>" = "\<كي داير>") due to the lack of standardization (e.g., diacritics, agglutinations, borrowings), making them overly rigid in cases where slight variations still convey the same meaning.
To gain a more fine-grained insight, we also included 
chrF \citep{popovic-2015-chrf}, operating at the level of character $n$-grams.
In addition, to capture higher-level semantic similarity, we also used BERTScore \citep{zhang2019bertscore}, with DarijaBERT as the reference model for summarization, and multilingual BERT\footnote{\url{https://hf.co/google-bert/bert-base-multilingual-cased}} for translation. These evaluations were conducted in a zero-shot setting using greedy decoding, and some in a few-shot setting.	
The number of few-shot examples was chosen based on relevant literature and standard practices.	

For summarization evaluation, we also employ the LLM-as-a-Judge approach \citep{zheng2023judging}, where a model judges the preferred summary between a reference and a generated one, based on predefined criteria. 
We report the win-rate, defined as the percentage of instances where the generated summary is chosen over the reference.	
Detailed information on the judge model, prompt, bias mitigation, and selection criteria is in Appendix \ref{app:details:judge}.
DarijaAlpacaEval employs the same approach as LLM-as-a-Judge, where we choose Jais-13B-Chat, the first Arabic-specialized LLM, as the reference.
For these two evaluations, we applied the default sampling-based decoding.

\smallskip

\noindent\textbf{Baseline models}.
We compared Atlas-Chat with instruction-tuned models from new Jais series (including the \texttt{-family} models trained from scratch and the \texttt{-adapted} ones based on LLaMA 2), along with AceGPT, LLaMA 3.1, 3.2, and Gemma 2 (our base model).
Given that Atlas-Chat features 2B, 9B, and 27B sizes, we extended our comparison to the closest larger-sized model above 27B when available, while included all smaller-sized ones.

\smallskip

\noindent\textbf{Zero-shot performance}.
The evaluation results in Table \ref{tab:results} demonstrate the exceptional performance of Atlas-Chat models across all Darija benchmarks.
Compared to baseline models with 7B or fewer parameters, Atlas-Chat-2B shows significantly superior zero-shot performance. Atlas-Chat-2B surpassed its closest competitor, Jais-family-6.7B-chat, by performance gaps of 5.05\% on DarijaMMLU, 2.40\% on DarijaHellaSwag, 2.11\% on Belebele\_Ary, 27.13\% on DarijaAlpacaEval, and 17.08\% on sentiment analysis. In translation and summarization tasks, Atlas-Chat-2B outperformed other models across all evaluation metrics.

The strong zero-shot performance of Atlas-Chat is further enhanced by the larger-sized Atlas-Chat-9B, which consistently outperforms other baseline models with parameters less than or equal to 13B, achieving the highest scores in 14 out of 16 metrics. Its strength is especially evident in translation as it leads in all three metrics, chrF, BLEU, and BERTScore, by a significant margin. Moreover, the model excels in DarijaMMLU, DarijaHellaSwag, Belebele\_Ary, DarijaAlpacaEval, and sentiment analysis, surpassing larger models like AceGPT-13B-chat and Jais-family-13B-Chat.

Our largest model, Atlas-Chat-27B, consistently outperforms competitors, including Jais-family-30B-8k-chat and Gemma-2-27B-It. In DarijaMMLU, DarijaHellaSwag, Belebele\_Ary, and DarijaAlpacaEval, it achieves zero-shot performance gaps of 10.07\%, 12.76\%, 1.51\%, and 10.00\%, respectively, over the highest-performing competitor. Similarly, in translation and summarization tasks, Atlas-Chat-27B demonstrates significant zero-shot performance advantages over its closest competitor, with substantial performance improvements over all evaluation metrics.

\smallskip

\noindent\textbf{Few-shot performance}.
Atlas-Chat demonstrated further improvements when moving from the zero-shot to the few-shot setting, with the effect being particularly pronounced for the 9B and 27B models, especially on the Belebele\_Ary benchmark. However, this enhancement in few-shot performance is not observed for the Atlas-Chat-2B model, despite consistently outperforming competitors.

\smallskip

\noindent\textbf{Further analysis}.
Although Atlas-Chat-27B showed the best overall performance, it was outperformed in the sentiment analysis task by smaller counterparts like Atlas-Chat-9B. We hypothesize that this discrepancy might be inherited from our base  models, where Gemma-2-9B-it similarly outperformed  Gemma-2-27B-it in the same task. 

Additionally, in the summarization task measured by ROUGE, Atlas-Chat models did not achieve a significant leading advantage as seen with other metrics. This discrepancy could stem from the inability of these $n$-gram-based metrics to fully capture Darija's nuances.	Moreover, summarization, as a less constrained generation task, often yields equally valid summaries that vary in formulation.		
However, when the models' summarization capability was evaluated using the LLM-as-a-judge framework, the judge model selected Atlas-Chat's responses 60.70\% of the time over reference summaries surpassing its closest competitor, Llama-3.1-8B-Instruct, by approximately 45\%.

\begin{figure}[t]
    \centering
    \includegraphics[width=1\linewidth]{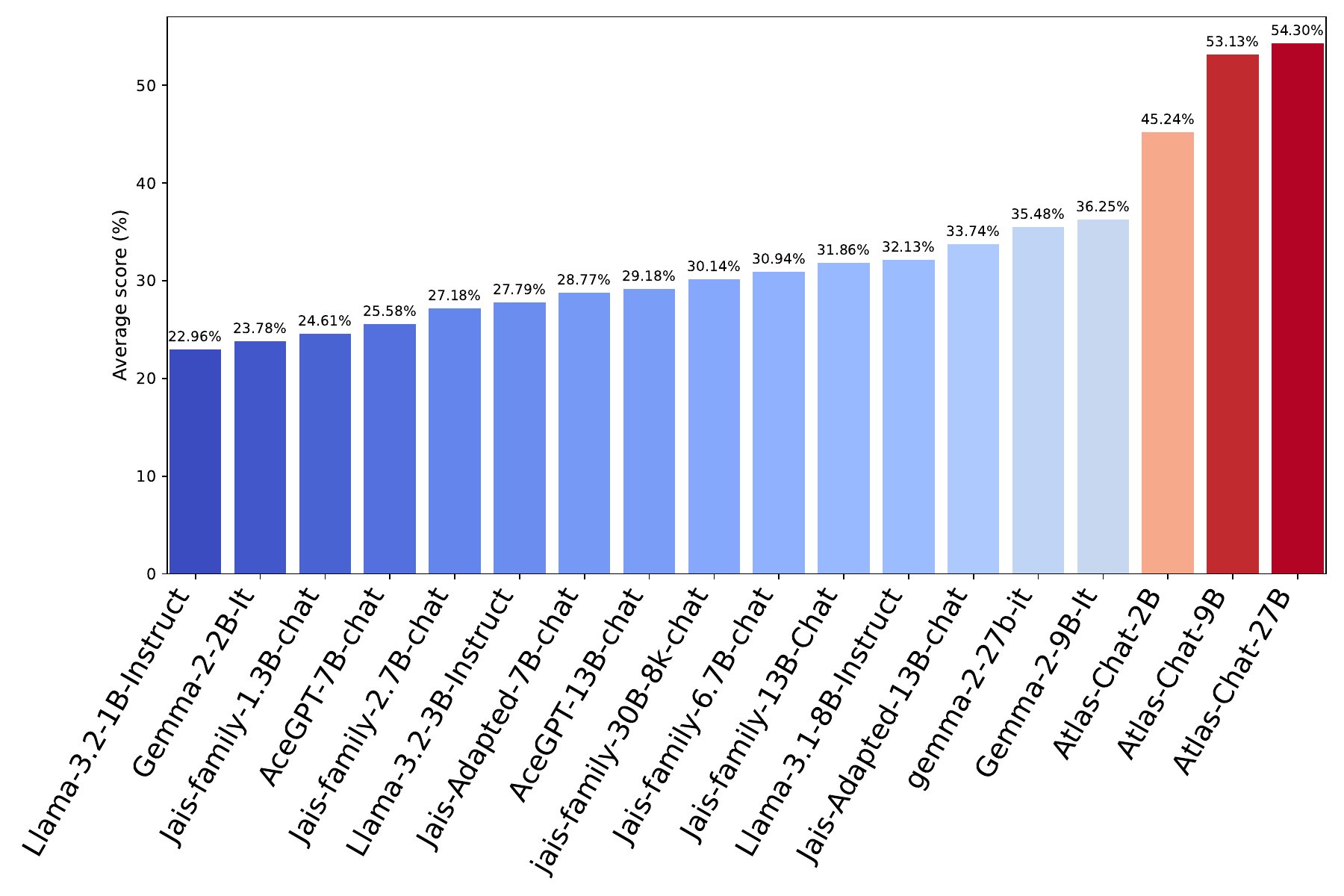}
    \caption{Average model scores over all  benchmarks.}
    \label{fig:overall-accuracy}
\end{figure}

Similarly, in the translation task measured by BLEU, baseline models demonstrated unexpectedly low performance. Quality analysis indicated that the low performance was due to their inability to consistently produce Darija. For example, in English-to-Darija translation, these models produced outputs consisting solely of MSA or a mix of MSA and Darija, resulting in a notable lack of overlapping $n$-grams with the reference text.

\section{Conclusion}

In this work, we presented Atlas-Chat, the first collection of large language models specifically developed for dialectal Arabic, with a primary focus on Moroccan Darija. We constructed a comprehensive instruction dataset by consolidating existing Darija resources, creating novel datasets both manually and synthetically, and translating English instructions with rigorous control measures. 
To evaluate LLM performance in Darija, we also introduced several benchmarks including both discriminative and generative tasks.
Atlas-Chat models showed superior performance in following Darija instructions and executing standard NLP tasks, outperforming both state-of-the-art and Arabic-specialized LLMs.
Our work highlights the potential of targeted LLM development for underrepresented languages and offers comprehensive design methodologies of instruction-tuning that can be applied to similar language adaptation challenges.

\section*{Limitations}
Despite the promising results, our work has some limitations. First, the model occasionally generates hallucinations. Second, the dataset may contain inherent biases that could affect the model’s fairness and representation.	Additionally, we relied heavily on Claude for translating English instructions into Darija. However, because Claude is primarily trained on English and reflects Western cultural values, it may not fully capture the unique nuances of Darija. Moreover, our models lack preference-tuning to better align with Darija speakers.	We intend to address these limitations in future work.

\section*{Acknowledgments}
The authors would like to thank all the Moroccan Darija speakers who warmly contributed to this work from its inception, assisting with data annotation and selection, and evaluating the quality of model outputs in their language.

\bibliography{googlescholar,aclanthology}

\appendix
\onecolumn

\section{Instruction Data Templates} 
\label{app:instruction_templates}
In this section, we list the instruction templates used for constructing our Darija-SFT-Mixture dataset. 
\subsection{Machine Translation} 
\label{app:instruction_templates:machine_translation}

\begin{center}
\begin{tabularx}{350pt}{X}
\toprule
\setcode{utf8}
\textbf{user}: \hfill \small$\backslash n$\small\textit{[source language text]}\small$\backslash n$ :\small\textit{[target language]}\setcode{utf8} \small\<لل> \textit{[source language]} \small\<ترجم من> \\
\textbf{assistant:} \small\textit{[target language text]}\\
\bottomrule
\end{tabularx}
\end{center}

\subsection{Transliteration} 
\label{app:instruction_templates:transliteration}
\begin{center}
\begin{tabularx}{350pt}{X}
\toprule
\setcode{utf8}
\textbf{user}: \hfill\small$\backslash n$\small\textit{[source language text]}\small$\backslash n$:\small[\textit{source language}] \small\<كتب هادشي بالحروف ديال>\\
\textbf{assistant:} \small\textit{[target language text]}\\
\bottomrule
\end{tabularx}
\end{center}

\subsection{Sentiment Analysis} 
\label{app:instruction_templates:sentiment_analysis}

\begin{center}
\begin{tabularx}{250pt}{X}
\toprule
\setcode{utf8}
\textbf{user}:\hfill \small$\backslash n$\small\<شنو هو الإحساس ديال هاد الجملة؟>\\
\hfill$\backslash n$\small \textit{[source text]}:\small\<العبارة>\\
\setcode{utf8}\hfill\small$\backslash n$:\small\<الإحتمالات>\\
\setcode{utf8}\hfill\small$\backslash n$\small\<سلبي>- \\
\setcode{utf8}\hfill\small$\backslash n$\small\<ايجابي> -\\
\textbf{assistant:} \small\textit{[target]}\\
\bottomrule
\end{tabularx}
\end{center}

\subsection{Automatic Summarization}
\label{app:instruction_templates:automatic_summarization}

\begin{center}
\begin{tabularx}{250pt}{X}
\toprule
\setcode{utf8}
\textbf{user}: \small\hfill$\backslash n$:\small\<لخص هاد المقطع>\\
\hfill \small$\backslash n$ \textit{[passage]}\\
\textbf{assistant:} \small\textit{[summary]}\\
\bottomrule
\end{tabularx}
\end{center}

\subsection{MoroccanWikipedia-QA}
\label{app:instruction_templates:moroccanwikipedia_QA}
\noindent Template 1:
\begin{center}
\begin{tabularx}{250pt}{X}
\toprule
\setcode{utf8}
\textbf{user}: \hfill\small$\backslash n\backslash n$:\small \<قرا هاد النص وجاوب على السؤال> \\
\hfill \small$\backslash n\backslash n$ \small\textit{[passage]}\\
\hfill \small$\backslash n\backslash n$ \small\textit{[question]}\\
\textbf{assistant:} \small \textit{[answer]}\\
\bottomrule
\end{tabularx}
\end{center}

\noindent Template 2:
\begin{center}
\begin{tabularx}{350pt}{X} 
\toprule
\textbf{user}: \hfill\small$\backslash n\backslash n$ \small\textit{[question]}\small $\backslash n$ :\small\<ودابا جاوب على هاد السؤال> $\backslash n\backslash n$\small\textit{[passage]} $\backslash n\backslash n$:\small\<قرا هاد النص>\\ 
\textbf{assistant:} \small\textit{[answer]}\\
\bottomrule
\end{tabularx}
\end{center}

\noindent Template 3:
\begin{center}
\begin{tabularx}{350pt}{X} 
\toprule
\setcode{utf8}
\textbf{user}:\hfill  $\backslash n\backslash n$\small\textit{[passage]} $\backslash n\backslash n$\small\<جاوب على هاد السؤال انطلاقا من داكشي لي فالنص>\\
\hfill \small $\backslash n\backslash n$ \small\textit{[question]}\\
\textbf{assistant:} \small \textit{[answer]}\\
\bottomrule
\end{tabularx}
\end{center}

\subsection{MoroccanSocialMedia-MultiGen}
\label{app:instruction_templates:moroccansocialmedia_multigen}

Continuation
\begin{center}
\begin{tabularx}{250pt}{X}
\toprule
\setcode{utf8}
\textbf{user}: \hfill \small$\backslash n$ \small\textit{[source sentence]} \small$\backslash n$ \small \<:كمل هاد الجملة>\\
\textbf{assistant:} \small\textit{[completion]}\\
\bottomrule
\end{tabularx}
\end{center}
\newpage
\noindent Reply
\begin{center}
\begin{tabularx}{250pt}{X}
\toprule
\setcode{utf8}
\textbf{user}: \hfill\small$\backslash n$ \textit{[message]}\small$\backslash n$ \small:\<جاوب على هاد الميساج>\\
\textbf{assistant:} \small \textit{[reply]}\\
\bottomrule
\end{tabularx}
\end{center}

\noindent Summarization
\begin{center}
\begin{tabularx}{250pt}{X}
\toprule
\setcode{utf8}
\textbf{user}: \hfill\small$\backslash n$ \textit{[passage]}\small$\backslash n$ \small:\<لخص هاد النص>\\
\textbf{assistant:} \small \textit{[summary]}\\
\bottomrule
\end{tabularx}
\end{center}

\noindent Rephrasing
\begin{center}
\begin{tabularx}{255pt}{X}
\toprule
\setcode{utf8}
\textbf{user}: \hfill\small$\backslash n$ \textit{[source sentence]}\small$\backslash n$ \small:\<كتب هاد الجملة بشي طريقة اخرى>\\
\textbf{assistant:} \small \textit{[resphrased sentence]}\\
\bottomrule
\end{tabularx}
\end{center}

\noindent Explanation
\begin{center}
\begin{tabularx}{250pt}{X}
\toprule
\setcode{utf8}
\textbf{user}: \hfill\small$\backslash n$ \textit{[source sentence]}\small$\backslash n$ \small:\<شرح ليا هاد الجملة>\\
\textbf{assistant:} \small \textit{[explanation]}\\
\bottomrule
\end{tabularx}
\end{center}

\noindent Safe Response
\begin{center}
\begin{tabularx}{250pt}{X}
\toprule
\setcode{utf8}
\textbf{user}: \hfill\small$\backslash n$ \textit{[source sentence]}\small$\backslash n$ \small:\<جاوب على هادشي بطريقة مأدبة>\\
\textbf{assistant:} \small \textit{[safe response]}\\
\bottomrule
\end{tabularx}
\end{center}

\subsection{DarijaStory-Completion}
\label{app:instruction_templates:darijastory_9esa}
\begin{center}
\begin{tabularx}{250pt}{X}
\toprule
\setcode{utf8}
\textbf{user}: \hfill\small$\backslash n$ \textit{[story]}\small$\backslash n$ \small:\<كمل هاد لقصة>\\
\textbf{assistant:} \small \textit{[completion]}\\
\bottomrule
\end{tabularx}
\end{center}

\bigskip

\section{TÜLU-V2-mix and Translation}
\label{app:tulu}
In this section, we provide a detailed overview of the TÜLU-V2-mix dataset and its translation process into Darija, including the datasets it incorporates and the sampling strategies employed. We also describe the dataset's format and the steps involved in translating the dataset to Moroccan Darija.
\subsection{Composition of TÜLU-V2-mix}
\label{app:tulu:composition}
TÜLU-V2-mix incorporates subsets from the following datasets: FLAN \citep{wei2021finetuned}\footnote{\url{https://github.com/google-research/FLAN/tree/main}}, Open Assistant 1 \citep{kopf2024openassistant}\footnote{\url{https://hf.co/datasets/OpenAssistant/oasst1}}, ShareGPT \citep{chen2023sharegpt4v}\footnote{\url{https://hf.co/datasets/anon8231489123/ShareGPT_Vicuna_unfiltered}}, GPT4-Alpaca \citep{peng2023instruction}\footnote{\url{https://github.com/Instruction-Tuning-with-GPT-4/GPT-4-LLM\#data-release}}, Code-Alpaca\footnote{\url{https://github.com/sahil280114/codealpaca}}, LIMA \citep{zhou2024lima}\footnote{\url{https://hf.co/datasets/GAIR/lima}}, WizardLM Evol Instruct \citep{xu2023wizardlm}\footnote{\url{https://hf.co/datasets/WizardLM/WizardLM_evol_instruct_V2_196k}}, and Open-Orca \citep{mukherjee2023orca}\footnote{\url{https://hf.co/datasets/Open-Orca/OpenOrca}}. The mixture also incorporates hard-coded instructions and a set of science-related questions derived from scientific documents. Table \ref{tab:tulu_subsets} presents descriptions of each of these datasets and describes how the subset in TÜLU-V2-mix was sampled.

\begin{table*}[ht]
\centering
\scriptsize  
\renewcommand{\arraystretch}{1.4}
\setlength{\tabcolsep}{2pt}  
\resizebox{\textwidth}{!}{%
\begin{tabular}{p{1.8cm} p{5.5cm} p{4cm}}  
\hline
\textbf{Dataset} & \textbf{Description} & \textbf{Sampling Strategy} \\
\hline
\textbf{FLAN} & A collection of datasets with tasks such as question answering, summarization, translation, and more. & 100,000 examples from FLAN v2, split equally between general tasks and the CoT subset. \\
\textbf{Open Assistant 1} & A human-annotated assistant-style conversation corpus. & Top-ranked paths in conversation trees. 7,708 examples. \\
\textbf{ShareGPT} & User-shared conversations with ChatGPT and GPT-4. & 114,046 samples from a processed ShareGPT dataset. \\
\textbf{GPT4-Alpaca} & GPT-4 generated responses to prompts from Alpaca. & 20,000 samples. \\
\textbf{Code-Alpaca} & Coding instruction-tuning data generated by text-davinci-003. & All 20,022 examples. \\
\textbf{LIMA} & Carefully selected data with a special focus on quality. & All 1,030 examples. \\
\textbf{WizardLM Evol Instruct} & Automatic evolution of instruction datasets, enhancing the complexity and diversity of instructions. & 30,000 examples. \\
\textbf{Open-Orca} & Augmented FLAN data with additional generated explanations. & 30,000 samples generated by GPT-4. \\
\textbf{Hardcoded} & Prompts ensuring the model correctly answers questions about its identity or creators. & 14 samples each repeated 10 times = 140 total samples. \\
\textbf{Science} & Scientific documents understanding tasks. & 7,544 examples. \\
\hline
\end{tabular}%
}
\caption{Subsets of TÜLU-V2-mix.}
\label{tab:tulu_subsets}
\end{table*}

\subsection{Dataset Format}
\label{app:tulu:format}

TÜLU-V2-mix is structured in a "messages" format commonly used for conversational datasets. Each interaction consists of a sequence of messages, where each message is represented as a JSON object with at least two key-value pairs:
\begin{itemize}
    \item \textbf{"role"}: Specifies the role of the participant in the conversation. Typically, this is either "user" (the person asking questions or giving prompts) or "assistant" (the model’s response).
    \item \textbf{"content"}: Contains the actual text of the message. This is where the question, instruction, or response is written.
\end{itemize}

\noindent Figure \ref{fig:sample_example} shows how samples from TÜLU-V2-mix are formatted.

\begin{figure}[ht]
\begin{lstlisting}
[
    {   
        "role": "user",
        "content": "What is the latest release version of the Lua programming language?" 
    },
    {
        "role": "assistant",
        "content": "The latest release of the Lua programming language is Lua 5.4.4, released in January 2022. The latest major version, Lua 5.4, was released in June  2020 and introduced new features such as a new generational mode for garbage collection, const variables, and to-be-closed variables."
    }
]
\end{lstlisting}
    \caption{A Sample from TÜLU-V2-mix.}
    \label{fig:sample_example}
\end{figure}

The "messages" format is particularly useful for training conversational models as it simulates multi-turn conversations by incorporating alternating roles between user and assistant messages. This format ensures a clear distinction between user inputs and the model's responses. Additionally, during fine-tuning, the loss function is applied specifically to messages with the role "assistant," to focus optimization on improving response generation. We applied this format to structure the whole training dataset. 

\subsection{Translation to Darija}
\label{app:tulu:translation}

\subsubsection{Preprocessing}

Before translating the dataset into Darija, we applied several filters to ensure that the translation meets our quality requirements:
\begin{itemize}
    \item \textbf{Excluding the Science subset:} We removed this part because the questions often involved parts or entire sections from research articles, which could lose meaning or coherence when translated, particularly into Darija. Additionally, we considered that a typical Darija-speaking user is unlikely to ask the model about research papers in Darija, as they would more commonly use English for such inquiries.
    \item \textbf{Filtering out empty messages:} Based on a reported issue\footnote{\url{https://github.com/allenai/open-instruct/issues/161}}, we discovered that some examples contained turns where the message role was defined, but the content was empty. To ensure data quality, we removed all such samples from the dataset.
    \item \textbf{Removing translation tasks:} We decided to omit translation instructions because translating both the source and target sentences into Darija would result in redundant outputs. Even if we specify that only the target sentence should be translated, it would be challenging to consistently ensure that the model performing the Darija translation adheres to the instruction across all examples. Additionally, verifying the quality of the translations would be challenging, particularly when the original meaning could be distorted. Furthermore, we already possess high-quality translation datasets, so including lower-quality translations would only degrade the overall dataset quality.\\
    To filter out translation tasks, we removed all samples containing either the strings "translate " or " translation ". We recognize that this method might exclude some instances where translation is mentioned without being the core task, for example, the user might be asking about the definition of the word "translation". However, given the large size of TÜLU-V2-mix, we believe such cases are rare, and the potential loss of a few samples would not impact the dataset's overall quality.
    \item \textbf{Excluding non-English samples:} We filtered out non-English examples to ensure higher translation quality, as translating from English to Darija tends to yield more accurate results compared to translations from other languages, especially those with low resources.\\
    To implement this filter, we used one of the best language identification tools: the fastText Language Identification model\footnote{\url{https://hf.co/facebook/fasttext-language-identification}}. We set \texttt{k=2}, meaning the model predicts the two most likely languages for each input text and provides a probability score for each. We excluded any samples where the most likely language was not English, as well as those labeled as English with a confidence score below 80\%. Through multiple experiments, we found that purely English texts typically score close to 100\%, while lower scores often indicate the presence of other languages mixed with English.
\end{itemize}

\subsubsection{Translation}

We experimented with several open-source and closed-source Darija translation models, including NLLB-200-3.3B\footnote{\url{https://hf.co/facebook/nllb-200-3.3B}} (No Language Left Behind\footnote{\url{https://ai.meta.com/research/no-language-left-behind}}), Terjman-Ultra\footnote{\url{https://hf.co/atlasia/Terjman-Ultra}}, GPT-4o\footnote{\url{https://openai.com/index/hello-gpt-4o}}, Claude 3 Opus\footnote{\url{https://www.anthropic.com/news/claude-3-family}}, and Claude 3.5 Sonnet\footnote{\url{https://www.anthropic.com/news/claude-3-5-sonnet}}. Our results showed that closed-source models consistently outperformed open-source alternatives, with GPT-4o and Claude 3.5 Sonnet taking the lead. We ultimately chose \textbf{Claude 3.5 Sonnet}, as it slightly outperformed GPT-4o and offered compatibility with Amazon Bedrock.

Table \ref{tab:translation_example} shows a comparison of an instruction translated to Darija using each of the models we tested. We observed that open-source models, namely NLLB-200-3.3B and Terjman-Ultra, tend to use more MSA, while closed-source models produce translations closer to Moroccan Darija. They also retain key formatting elements like line breaks (\textbackslash n) and tags (\#\#\#), which are crucial for preserving the structure of the instructions.

\begin{table}[ht]
\small
\centering
\renewcommand{\arraystretch}{1.4}
\begin{tabular}{p{1.6cm}  p{12.4cm}}
\hline
\textbf{Original Sentence} & \texttt{Write a response that appropriately completes the request.\textbackslash n\textbackslash n\#\#\# Instruction:\textbackslash nIdentify four positive impacts that artificial intelligence can have on the healthcare industry\textbackslash n\textbackslash n\#\#\# Response:}
\\
\hline
\textbf{NLLB-200-3.3B} &  \setcode{utf8}\small{\begin{arabtext}
كتب رد يكمل الطلب بشكل مناسب. \#\#\# التعليمات: حدد أربعة من التأثيرات الإيجابية التي يمكن لذكاء اصطناعي أن يكون لها على صناعة الرعاية الصحية \#\#\# الرد:
\end{arabtext}}\\
\textbf{Terjman-Ultra} & \setcode{utf8}\small{\begin{arabtext}
كتبي رد ياكمل الطلب بشكل مناسب. \#\#\# تعليمات: حدد أربعة تأثيرات إيجابية يمكن لذكاء اصطناعي أن يكون لها على صناعة الرعاية الصحية \#\#\# رد:
\end{arabtext}} \\
\textbf{GPT-4o} & \setcode{utf8}\small{\begin{arabtext}
كتب رد كيكمّل الطّلب بشكل مناسب.$\backslash n$$\backslash n$\#\#\# التعليمات:$\backslash n$حدّد أربع تأثيرات إيجابية اللي الذكاء الاصطناعي يمكن يكون عندو على قطاع الرعاية الصحية$\backslash n$$\backslash n$\#\#\# الرد:
\end{arabtext}} \\
\textbf{Claude 3 Opus} & \setcode{utf8}\small{\begin{arabtext}
أكتب جواب لي يكمل الطلب بشكل مناسب.$\backslash n$$\backslash n$\#\#\# التعليمات:$\backslash n$عرف أربع تأثيرات إيجابية لي يمكن للذكاء الاصطناعي يكون عندو على قطاع الصحة$\backslash n$$\backslash n$\#\#\# الجواب:
\end{arabtext}} \\
\textbf{Claude 3.5} 

\textbf{Sonnet} & \setcode{utf8}\small{\begin{arabtext}
اكتب جواب اللي يكمل الطلب بشكل مناسب.$\backslash n$$\backslash n$\#\#\# التعليمات:$\backslash n$حدد أربعة تأثيرات إيجابية اللي يمكن للذكاء الاصطناعي يديرها على قطاع الصحة$\backslash n$$\backslash n$\#\#\# الجواب:
\end{arabtext}} \\
\hline
\end{tabular}
\caption{Translation example for model comparison.}
\label{tab:translation_example}
\end{table}

We used Amazon Bedrock\footnote{\url{https://aws.amazon.com/bedrock}}, a cloud-based machine learning service from AWS, to translate the dataset into Darija. We provided specific instructions to Claude 3.5 Sonnet for handling the translations, refining the prompt after several rounds of experimentation. The final version of the prompt that produced the best results is shown in Figure \ref{fig:translation_prompt}. We altered this prompt slightly as needed for each subset of the dataset, ensuring that the translation remained consistent with the context and structure of each specific subset.

\begin{figure*}[ht]
    \raggedleft
\begin{lstlisting}
Translate the 'content' field in the paragraph after [Source Text] to Moroccan dialect (Darija - Arabic alphabet) while following these guidelines:
                
- Keep the format of the original text (list of json).
- If a word is usually not used in Arabic, use its French equivalent.
- Do not include any introduction or explanation after the translation, only the translation.
- If there is a given context, example or question translate it as well.
- Whenever you come across code contexts or technical words, keep them in English.
- Whenever you come across literature, or example or question, translate it to Moroccan.
- If the text is culturaly not accepted for Morrocans, change it to a more acceptable one.
- Do not answer the request in the source text.
- Write first the original text after the tag [[Original]] and then the translation after the tag [[Translation]].
    
[Source Text]

\end{lstlisting}
    
    \caption{The prompt given to Claude 3.5 Sonnet for translation.}
    \label{fig:translation_prompt}
\end{figure*}

We used asynchronous programming techniques with Python’s asyncio library to make multiple concurrent requests (often 25 at a time) to the Bedrock translation service. This parallel approach significantly sped up the overall translation process by allowing us to handle many requests simultaneously instead of sequentially.

\subsubsection{Postprocessing}

After finishing the translation and cleaning the errors, we post-processed the translations by:
\begin{itemize}
    \item \textbf{Replacing non-translated keywords:} Some keywords such as "Input", "Output", "Response", "Answer", "Instructions", "Hypothesis" and "Additional Context" were not translated. We replaced these keywords with their Darija equivalents: \< المعطيات, النتيجة, الجواب, الجواب, تعليمات, الفرضية, سياق إضافي.>
    \normalsize
    \item \textbf{Removing samples with excessive English content:} We utilized the fastText Language Identification model to detect samples where the predicted language was not Arabic. Since the model does not differentiate dialects, Darija is recognized as Arabic due to its use of Arabic script. We removed samples where the predicted language was not Arabic or where Arabic was predicted with a confidence level below 80\%.    
\end{itemize}

\bigskip

\section{Additional Details}
\subsection{Hard Coded Instruction Samples}
\label{app:details:hard_coded}
\begin{figure}[H]
    \centering
    \includegraphics[width=\textwidth]{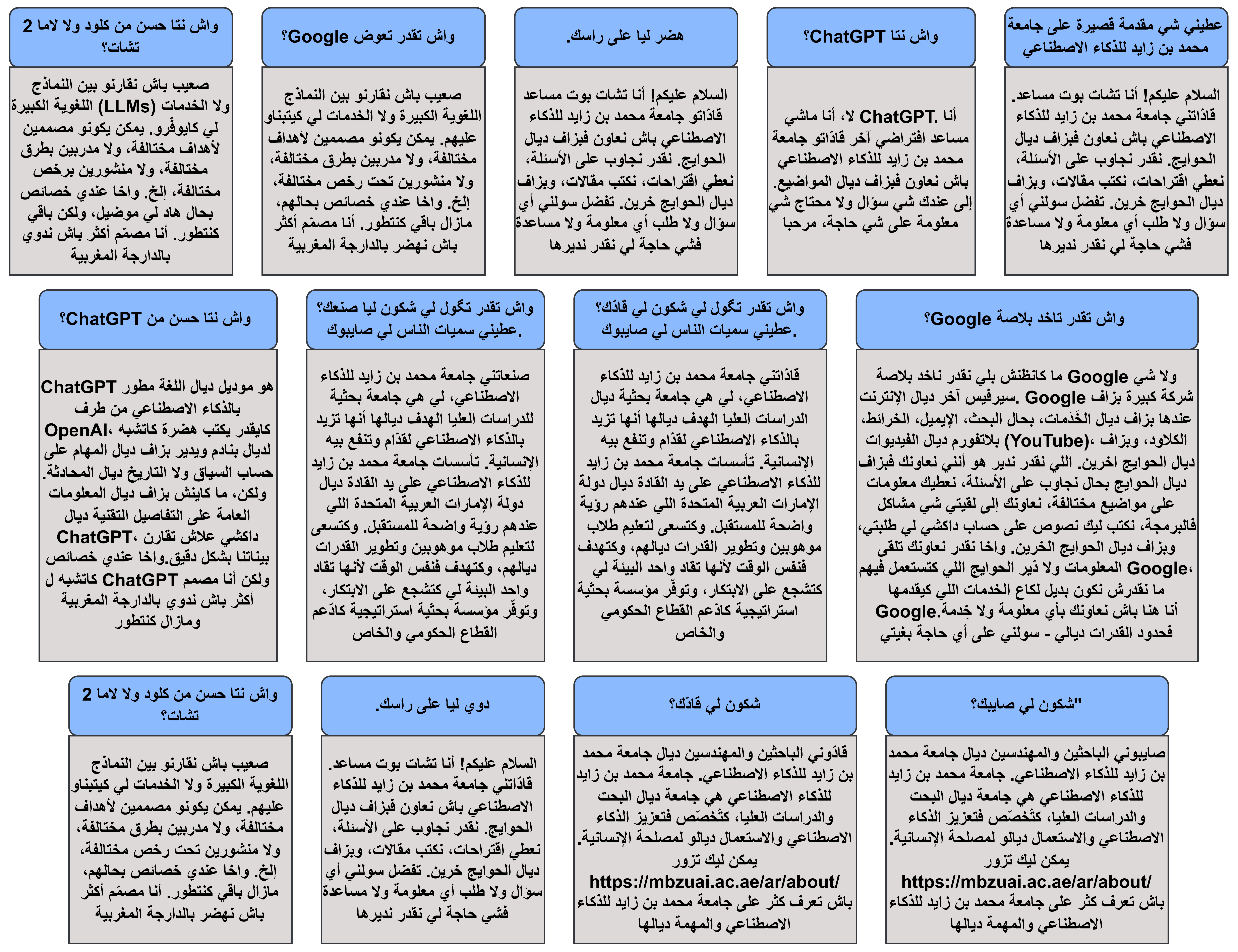}
    \caption{Hard coded instruction-answer pairs.}
    \label{fig:Atlas-Chat_hard_coded}
\end{figure}

We manually created 13 instruction samples to ensure that the model responds correctly to identity and creator-related questions, such as ``Who created you?'' and ``What is your name?''. Each instruction is repeated 10 times to reinforce the memorization of the answers. Figure \ref{fig:Atlas-Chat_hard_coded} presents the full list of hard-coded instruction-answer pairs.\vspace{-10pt}

\newpage

\subsection{Selected Keywords for Tweet Searching}
\label{app:details:keywords}
We provide the 26 Darija-specific keywords used for tweet collection through the Twitter API, as referenced in Section \ref{sec:data:msm_mg}.
\begin{arabtext}كاتشوف, كيضحك, كتبكي, داكشي, كيشوف, كتشوف, كيزيدو, دابا, ديال, تبوڭيصة, مكلخ, حشومة, منبقاوش, شلاهبية, تخربيق, كايدوي, كاندوي, يسيفطوه, يصيفطوه, السماسرية, ماكينش, مزيانين, الفقصة, زوينين, سيمانة, الدراري.
\end{arabtext}
\normalsize
\subsection{DarijaAlpacaEval Dataset Creation and Models Evaluation}
\label{app:details:alpaca}
To create the DarijaAlpacaEval dataset, we employed Claude 3.5 Sonnet to translate and culturally adapt the AlpacaEval dataset \cite{alpaca_eval} for evaluating models' capabilities in instruction following in Moroccan Darija. The prompt used for translation is shown in Figure \ref{fig:alpaca_translation_prompt}.

\begin{figure*}[ht]
    \raggedleft
\begin{lstlisting}
Given the following question about U.S. culture:{english_question}, translate and adapt it to focus on Moroccan culture. 
Ensure that the question retains the same underlying theme but is contextually suitable for Morocco, taking into account cultural, historical, and societal differences. 
For example, replace references to American holidays, traditions, or figures with their Moroccan counterparts.
The questions should be precise and should not differ significantly in length from the original question. 
Ensure that the question is unique to Morocco and not applicable to any neighboring countries. 
Adjust the language from English to Arabic Moroccan Darija.
Return only the question with no additional text.
\end{lstlisting}
    
\caption{The prompt given to Claude 3.5 Sonnet for translation and cultural adaptation of the AlpacaEval instructions.}
\label{fig:alpaca_translation_prompt}
\end{figure*}

This process resulted in 805 instructions, all adapted to the Moroccan culture and written in Darija. The models were subsequently evaluated by generating responses to these instructions, with their answers compared to a baseline model, jais-13b-chat, one of the earliest state-of-the-art models developed for Arabic NLP tasks. To assess cultural appropriateness, Claude 3.5 Sonnet was prompted to compare two model responses for each instruction, using criteria focused on cultural alignment, fluency, and relevance. The evaluation prompt is show in Figure \ref{fig:alpaca_evaluation_prompt}.

\begin{figure*}[ht]
    \raggedleft
\begin{lstlisting}
You are an expert evaluator tasked with judging the cultural appropriateness and relevance of two answers written in Moroccan Darija for a given instruction. Your judgment should focus solely on how well the answers reflect Moroccan cultural norms, values, and context.

### Criteria:
1. Cultural Appropriateness and Relevance: The answer should align well with Moroccan culture, norms, and societal context. Avoid any references, language, or ideas that are not relevant or appropriate for Morocco.
2. Fluency: The answer has to be in clear and precise language in Moroccan Darija.
3. Relevance: The answer should answer the instruction without any divergence from the instruction's goal.

### Instructions:
For each instruction, you will receive two answers, A and B. Evaluate them based on the criterion above and decide which one better reflects Moroccan culture. Provide only the letter A or B as the answer.

### Output format:
Better Answer: [A or B]

### Evaluate:
**Instruction**: 
[Start of the instruction]
{instruction}
[Text of the instruction]

**Answer A**: 
[Start of Answer A]
{answer_a}
[Text of Answer A]

**Answer B**:
[Start of Answer B]
{answer_b}
[Text of Answer B]

Your Response (Only "A" or "B" with no additional text):
\end{lstlisting}
\caption{The prompt Given to Claude 3.5 Sonnet for choosing the answer that better follows the instruction and predefined DarijaAlpacaEval criteria between the baseline another LLMs generated answers.}
\label{fig:alpaca_evaluation_prompt}
\end{figure*}
Each pair of baseline and model answers, with positions swapped, was evaluated twice by Claude to determine the better answer. If the position swap influenced Claude's choice, that particular pair was discarded to ensure the method's rhobustness to possible LLM biases. The model’s win-rate was then calculated as the proportion of instances where Claude selected the model's answer over the baseline.
\subsection{Selected Topics from MMLU and ArabicMMLU}
\label{app:details:mmlu}
The \textbf{MMLU} subjects included in DarijaMMLU are: Global Facts, High School European History, High School Geography, High School Government and Politics, High School Psychology, High School Statistics, High School World History, Human Aging, International Law, Jurisprudence, Logical Fallacies, Management, Marketing, Moral Disputes, Moral Scenarios, Nutrition, Philosophy, Professional Law, Professional Psychology, Public Relations, Security Studies, Sociology, and World Religions. 

\smallskip

\noindent From \textbf{ArabicMMLU}, the subjects adopted into DarijaMMLU are: Islamic Studies, Driving Test, Natural Science, History, General Knowledge, Law, Physics, Social Science, Management, Arabic Language, Political Science, Philosophy, Accounting, Computer Science, Geography, Mathematics, Biology, Economics, Arabic Language (General), Arabic Language (Grammar), and Civics.

\subsection{LLM-as-a-Judge Prompt for Summarization Evaluation}
\label{app:details:judge}
Following the work of \citet{zheng2023judging} and \citet{fabbri-etal-2021-summeval}, which used advanced LLMs to evaluate responses from other LLMs, we employed Claude 3.5 Sonnet to assess the models' summarization capabilities.
Summarization is subjective, and traditional text overlap-based methods often struggle to provide accurate evaluations.	
As shown in Figure \ref{fig:llm_judge_prompt}, we instructed Claude to evaluate model-generated summaries based on three main criteria: wordness, conciseness, and relevance.
The objective of the Darija summarization task is to produce a concise summary in native Darija using the fewest words possible, without introducing external information.

At each evaluation step, two summaries were presented to Claude: one generated by an LLM and the corresponding ground truth summary.	
To mitigate biases such as verbosity and position bias, identified by \citet{zheng2023judging}, all models were instructed to generate summaries of no more than 30 words (the average length of title summaries).
Additionally, each pair of generated and ground truth summaries was presented to Claude twice, with their positions swapped.	
Pairs in which position swapping influenced Claude's decision were discarded.
The win-rate of a model's summary was calculated based on how often Claude preferred the model's summary over the ground truth.	

\begin{figure*}[ht]
    \raggedleft
    \begin{lstlisting}
You are an expert evaluator tasked with judging the quality of two summaries written in Moroccan Darija for a given passage, also in Moroccan Darija. You are strict regarding any language or dialect that is not Moroccan Darija, such as Modern Standard Arabic (MSA) and English. 

### Criteria:
Choose the better summary based on these criteria:
1. **Wordness**: Clear and precise language in Moroccan Darija that conveys the passage's original meaning and doesn't use any other language or Dialect.
2. **Conciseness**: Straight to the point, capturing essential information without unnecessary details.
3. **Relevance**: Directly related to the passage without adding new information.

### Instructions:
For each passage, you will receive two summaries, **A** and **B**. Evaluate them based on the criteria above and decide which one is better. Provide only the letter **A** or **B** as the answer.
It is strictly forbidden that a summary is written in Modern Standard Arabic (MSA). 
A summary should not be chosen if it is written in MSA.s

###Output format: 
Better Summary: [A or B]

### Evaluate:
**Passage**: 
[Start of the passage]
{passage}
[Text of the passage]

**Summary A**: 
[Start of Summary A]
{summary_a}
[Text of Summary A]

**Summary B**:
[Start of Summary B]
{summary_b}
[Text of Summary B]

Your Response (Only A or B with no additional text):
    \end{lstlisting}
    
    \caption{The prompt Given to Claude 3.5 Sonnet for choosing the best summary between the baseline and LLM-generated summaries.}
    \label{fig:llm_judge_prompt}
\end{figure*}

\bigskip
\newpage

\section{Examples of Atlas-Chat-9B Responses}
\begin{figure}[ht]
    \centering
    \includegraphics[width=0.8\linewidth]{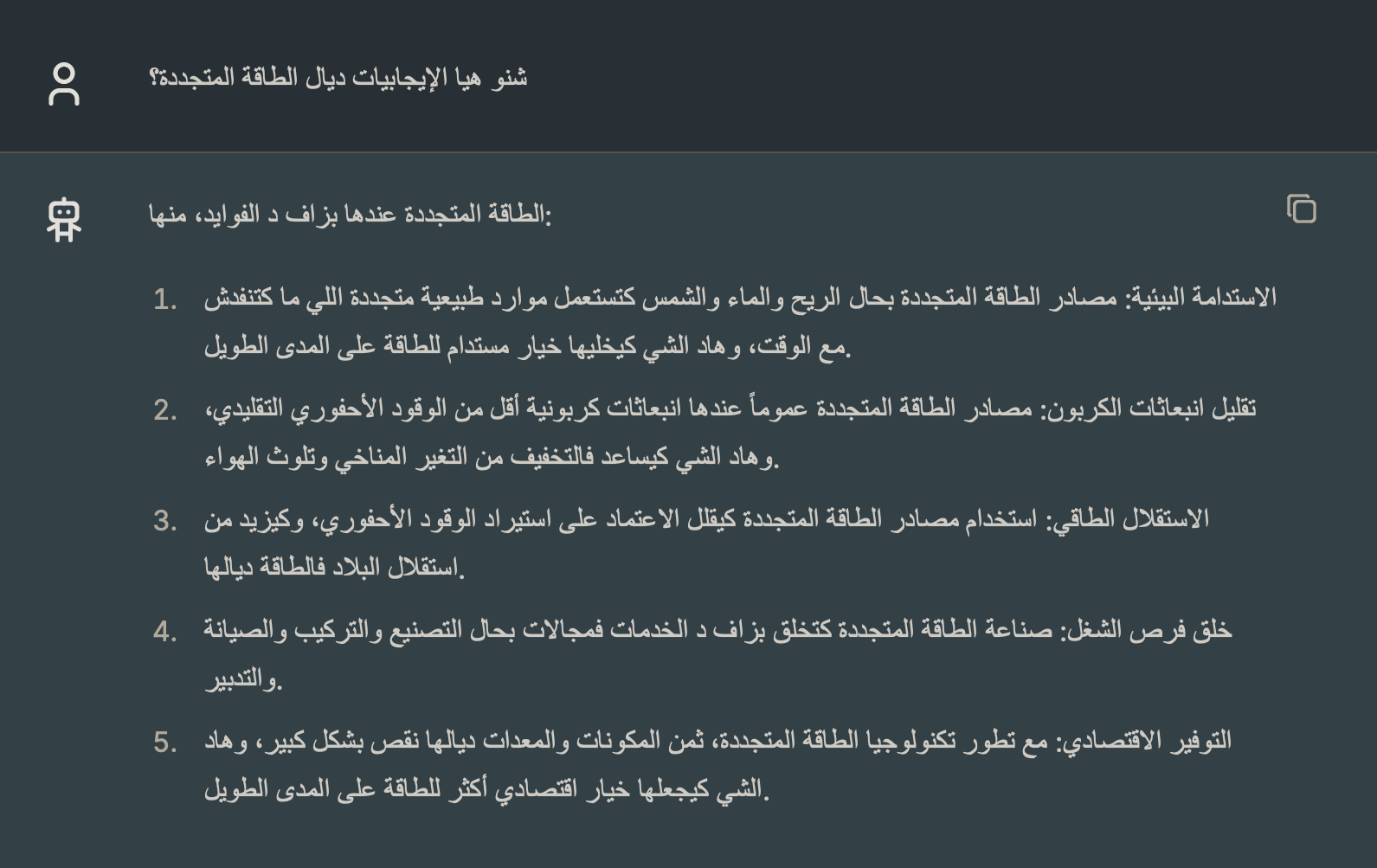}
    \includegraphics[width=0.8\linewidth]{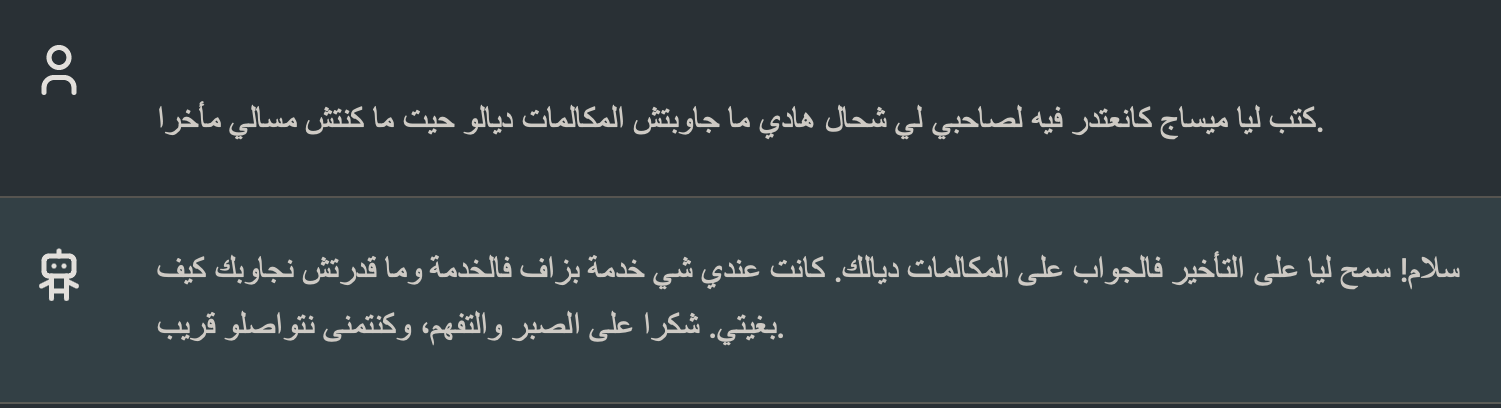}
    \includegraphics[width=0.8\linewidth]{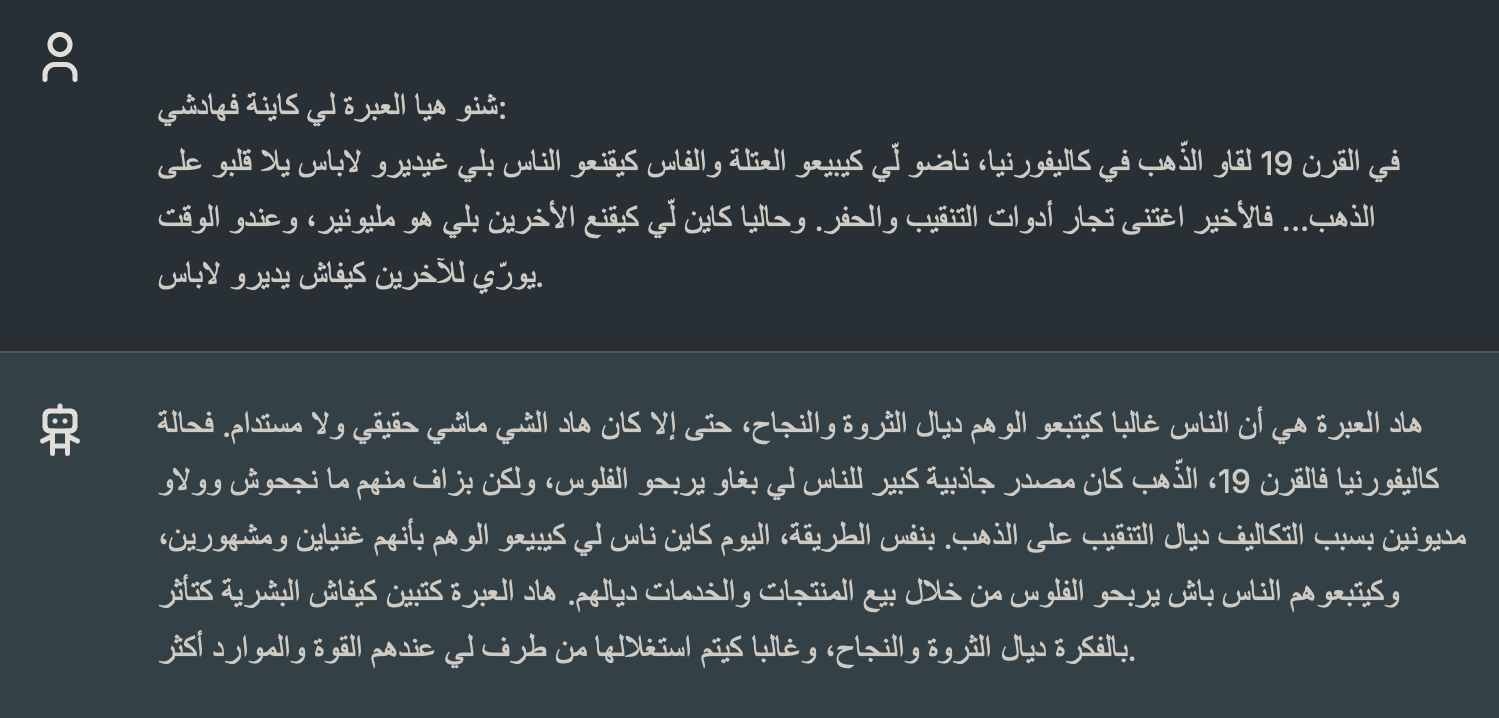}
    \caption{Atlas-Chat-9B response example 1.}
    \label{fig:response_eg_1}
\end{figure}

\begin{figure}[ht]
    \centering
    \includegraphics[width=0.8\linewidth]{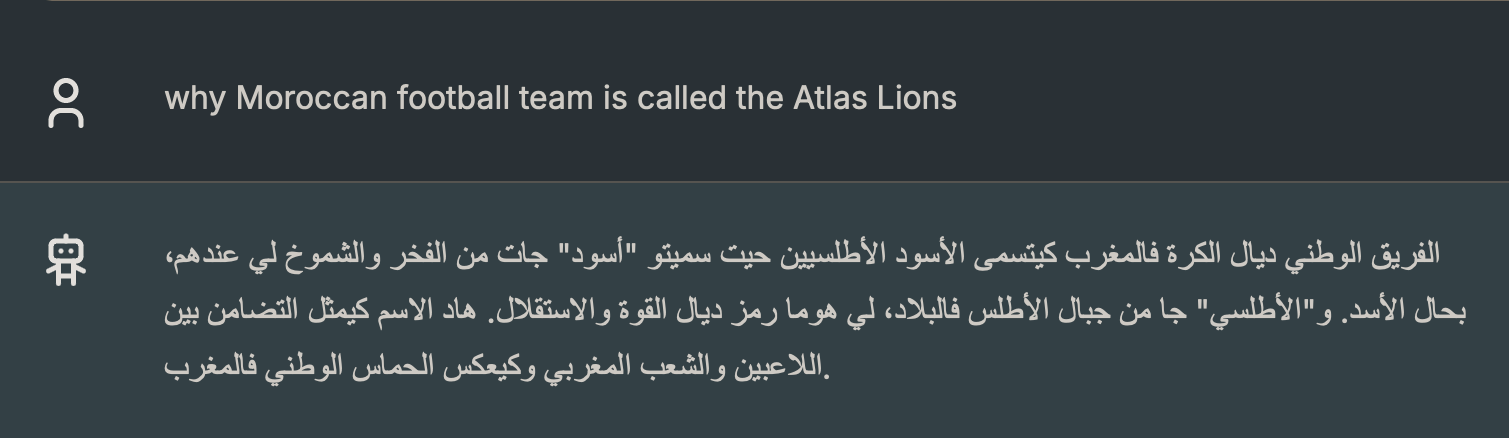}
    \caption{\centering Atlas-Chat-9B response example 2 (The model can understand English instructions but only responds in Darija).}
    \label{fig:response_eg_2}
\end{figure}

Figure \ref{fig:response_eg_1} and \ref{fig:response_eg_2} present some samples of Atlas-Chat responses on a variety of questions.
\label{app:chat_examples}

\end{document}